\pdfoutput=1

\documentclass[11pt]{article}

\usepackage{acl}
\usepackage{amsmath,amstext}
\usepackage{amsfonts,amssymb}
\usepackage{booktabs}
\usepackage{multirow}
\usepackage{graphicx}
\makeatletter
\renewcommand{\maketag@@@}[1]{\hbox{\m@th\normalsize\normalfont#1}}%
\makeatother

\usepackage{times}
\usepackage{latexsym}

\usepackage[T1]{fontenc}

\usepackage[utf8]{inputenc}

\usepackage{microtype}

\title{Denoising Bottleneck with Mutual Information Maximization \\for Video Multimodal Fusion}

\author{
\textbf{{Shaoxiang Wu}\textsuperscript{\rm1}, {Damai Dai}\textsuperscript{\rm 1}, {Ziwei Qin}\textsuperscript{\rm 1}, {Tianyu Liu}\textsuperscript{\rm 2},} \\
\textbf{ {Binghuai Lin}\textsuperscript{\rm 2}, {Yunbo Cao}\textsuperscript{\rm 2}, {Zhifang Sui}\textsuperscript{\rm 1}} \\
\textsuperscript{\rm 1}MOE Key Lab of Computational Linguistics, Peking University\\
\textsuperscript{\rm 2}Tencent Cloud AI
\\
{\tt wushaoxiang@stu.pku.edu.cn} ~~ {\tt \{daidamai,szf\}@pku.edu.cn}
}

\begin{document}
\maketitle
\begin{abstract}
Video multimodal fusion aims to integrate multimodal signals in videos, such as visual, audio and text, to make a complementary prediction with multiple modalities contents.
However, unlike other image-text multimodal tasks, video has longer multimodal sequences with more redundancy and noise in both visual and audio modalities.
Prior denoising methods like forget gate are coarse in the granularity of noise filtering. 
They often suppress the redundant and noisy information at the risk of losing critical information.
Therefore, we propose a denoising bottleneck fusion (DBF) model for fine-grained video multimodal fusion. 
On the one hand, we employ a bottleneck mechanism to filter out noise and redundancy with a restrained receptive field. 
On the other hand, we use a mutual information maximization module to regulate the filter-out module to preserve key information within different modalities.
Our DBF model achieves significant improvement over current state-of-the-art baselines on multiple benchmarks covering multimodal sentiment analysis and multimodal summarization tasks. 
It proves that our model can effectively capture salient features from noisy and redundant video, audio, and text inputs.
The code for this paper is publicly available at \url{https://github.com/WSXRHFG/DBF}. 
\end{abstract}

\section{Introduction}

\begin{figure*}[!ht]
    \centering
    \includegraphics[width=\linewidth]{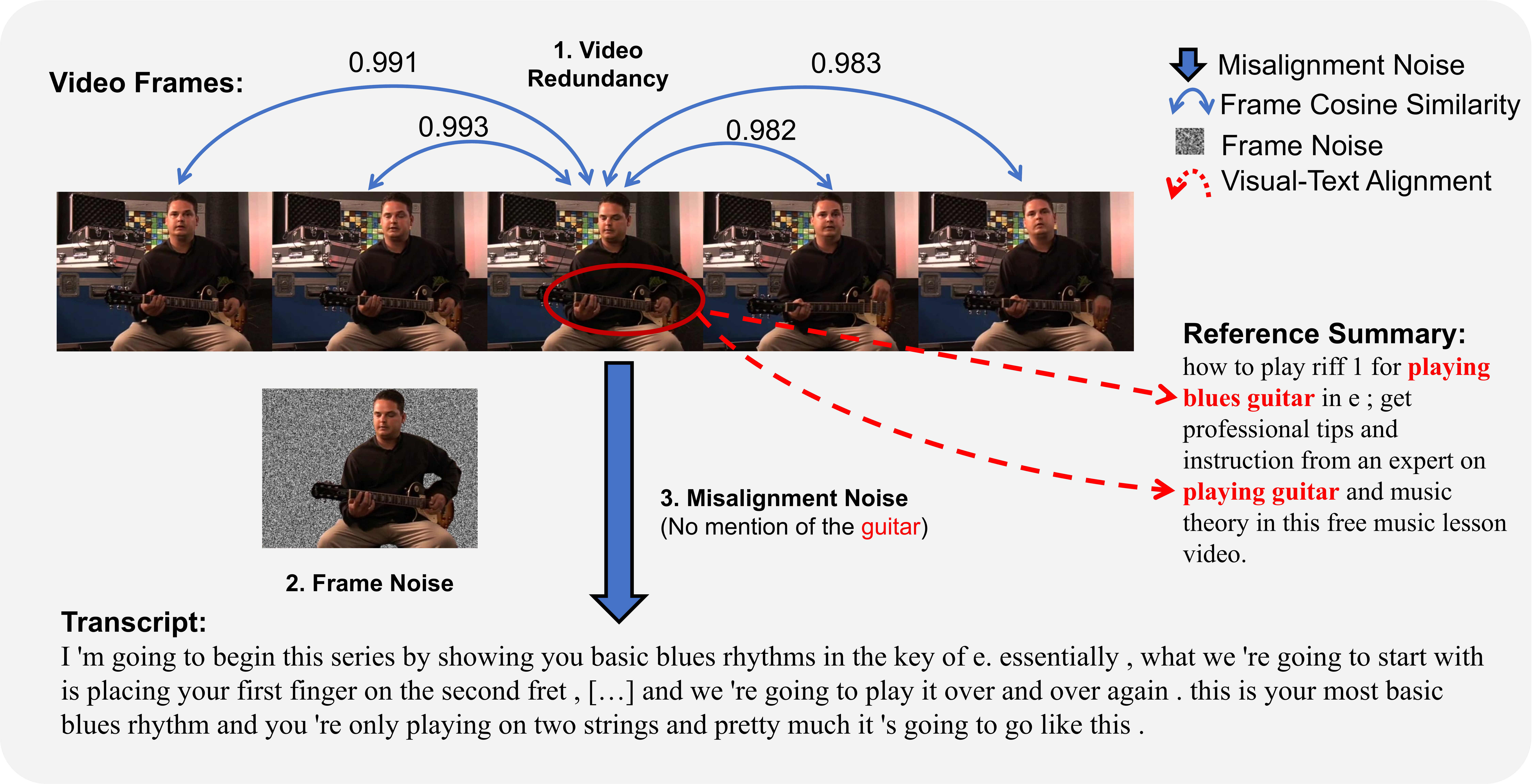}
    \caption{
    An example of redundancy and noise in a video. 
    As illustrated, consecutive frames have high cosine similarity, which results in a problem of \textbf{redundancy}.  
    In addition, useless information like distracting background and weak alignment between frames and transcripts compose \textbf{noises} in videos.
    }
    \label{fig:example}	
\end{figure*}

With the rapid development of social platforms and digital devices, more and more videos are flooding our lives, which leads video multimodal fusion an increasingly popular focus of NLP research.
Video multimodal fusion aims to integrate the information from two or more modalities (e.g., visual and audio signals) into text for a more comprehensive reasoning. 
For example, multimodal sentiment analysis \citep{poria2020beneath} utilizes contrast between transcript and expression to detect sarcam, multimodal summarization \citep{sanabria2018how2} complete summary with information only exists in visual signal. 

However, as shown in the Figure \ref{fig:example}, there exist plenty of redundancy and noise in video multimodal fusion:
1) high similarity across consecutive frames brings \emph{video redundancy}; 
2) useless information, such as the distracting background, introduces \emph{frame noise};
3) weak alignment between visual stream and text also introduces \emph{misalignment noise}.
To alleviate the problem of redundancy and noise in video multimodal fusion, \citet{liu2020multistage} control the flow of redundant and noisy information between multimodal sequences by a fusion forget gate.
The fusion forget gate impairs the impact of noise and redundancy in a coarse grain of the whole modality, so it will also filter out some representative information in the filtered modality. 

In order to remove noise and redundancy while preserving critical information in video multimodal fusion, we propose a denoising fusion bottleneck (DBF) model with mutual information maximization (MI-Max).
Firstly, inspired by \citet{nagrani2021attention}, we introduce a bottleneck module to restrict the redundant and noisy information across different modalities. 
With the bottleneck module, inputs can only attend to low-capacity bottleneck embeddings to exchange information across different modalities, which urges redundant and noisy information to be discarded. 
Secondly, in order to prevent key information from being filtered out, we adopt the idea of contrastive learning to supervise the learning of our bottleneck module. 
Specifically, under the noise-contrastive estimation framework \citep{gutmann2010noise}, for each sample, we treat all the other samples in the same batch as negative ones. 
Then, we aim to maximize the mutual information between fusion results and each unimodal inputs by distinguishing their similarity scores from negative samples. 
Two aforementioned modules complement each other, the MI-Max module supervises the fusion bottleneck not to filter out key information, and in turn, the bottleneck reduces irrelevant information in fusion results to facilitate the maximization of mutual information. 

We conduct extensive experiments on three benchmarks spanning two tasks. 
MOSI \citep{zadeh2016mosi} and MOSEI \citep{mosei} are two datasets for multimodal sentiment analysis. 
How2 \citep{sanabria2018how2} is a benchmark for multimodal summarization. 
Experimental results show that our model achieves consistent improvements compared with current state-of-the-art methods. 
Meanwhile, we perform comprehensive ablation experiments to demonstrate the effectiveness of each module. 
In addition, we visualize the attention regions and tensity to multiple frames to intuitively show the behavior of our model to reduce noise while retaining key information implicitly. 

Concretely, we make the following contributions: 
(i) We propose a denoising bottleneck fusion model for video multimodal fusion, which reduces redundancy and noise while retaining key information. 
(ii) We achieve new state-of-the-art performance on three benchmarks spanning two video multimodal fusion tasks. 
(iii) We provide comprehensive ablation studies and qualitative visualization examples to demonstrate the effectiveness of both bottleneck and MI-Max modules.

\section{Related Work}

We briefly overview related work about multimodal fusion and specific multimodal fusion tasks including multimodal summarization and multimodal sentiment analysis.

\subsection{Video Multimodal Fusion}

Video multimodal fusion aims to join and comprehend information from two or more modalities in videos to make a comprehensive prediction. 
Early fusion model adopted simple network architectures. \citet{zadeh2017tensor,liu2018efficient} fuse features by matrix operations;
and \citet{zadeh2018memory} designed a LSTM-based model to capture both temporal and inter-modal interactions for better fusion.
More recently, models influenced by prevalence of Transformer \citep{vaswani2017attention} have emerged constantly: \citet{zhang2019neural} injected visual information in the decoder of Transformer by cross attention mechanism to do multimodal translation task;
\citet{wu2021text} proposed a text-centric multimodal fusion shared private framework for multimodal fusion, which consists of the cross-modal prediction and sentiment regression parts.  
And now vision-and-language pre-training has become a promising practice to tackle video multimodal fusion tasks.
\citep{Sun_2019_ICCV} firstly extend the Transformer structure to video-language pretraining and used three pre-training tasks: masked language prediction, video text matching, masked video prediction.

In contrast to existing works, we focus on the fundamental characteristic of video: audio and visual inputs in video are redundant and noisy~\citep{nagrani2021attention} so we aim to remove noise and redundancy while preserving critical information. 

\subsection{Video Multimodal Summarization}

Video multimodal summarization aims to generate summaries from visual features and corresponding transcripts in videos. 
In contrast to unimodal summarization, some information~(e.g., guitar) only exists in the visual modality. 
Thus, for videos, utilization of both visual and text features is necessary to generate a more comprehensive summary. 

For datasets, \citet{li2017multi} introduced a multimodal summarization dataset consisting of 500 videos of news articles in Chinese and English. \citet{sanabria2018how2} proposed the How2 dataset consists of 2,000 hours of short instructional videos, each coming with a summary of two to three sentences.

For models, \citet{liu2020multistage} proposed a multistage fusion network with a fusion forget gate module, which controls the flow of redundant information between multimodal long sequences. Meanwhile, \citet{yu2021vision} firstly introduced pre-trained language models into multimodal summarization task and experimented with the optimal injection layer of visual features.

We also reduce redundancy in video like in \citep{yu2021vision}.
However, we do not impair the impact of noise and redundancy in a coarse grain with forget gate.
Instead, we combine fusion bottleneck and MI-Max modules to filter out noise while preserving key information.

\subsection{Multimodal Sentiment Analysis}

Multimodal sentiment analysis (MSA) aims to integrate multimodal resources, such as textual, visual, and acoustic information in videos to predict varied human emotions. 
In contrast to unimodal sentiment analysis, utterance in the real situation sometimes contains sarcasm, which makes it hard to make accurate prediction by a single modality.
In addition, information such as expression in vision and tone in acoustic help assist sentiment prediction.
\citet{yu2021learning} introduced a multi-label training scheme that generates extra unimodal labels for each modality and concurrently trained with the main task.
\citet{han2021improving} build up a hierarchical mutual information maximization guided model to improve the fusion outcome as well as the performance in the downstream multimodal sentiment analysis task.
\citet{luo2021scalevlad} propose a multi-scale fusion method to align different granularity information from multiple modalities in multimodal sentiment analysis.

Our work is fundamentally different from the above work. 
We do not focus on complex fusion mechanisms, but take the perspective of information in videos, and stress the importance of validity of information within fusion results. 

\begin{figure*}[!ht]
    \centering
    \includegraphics[width=\linewidth]{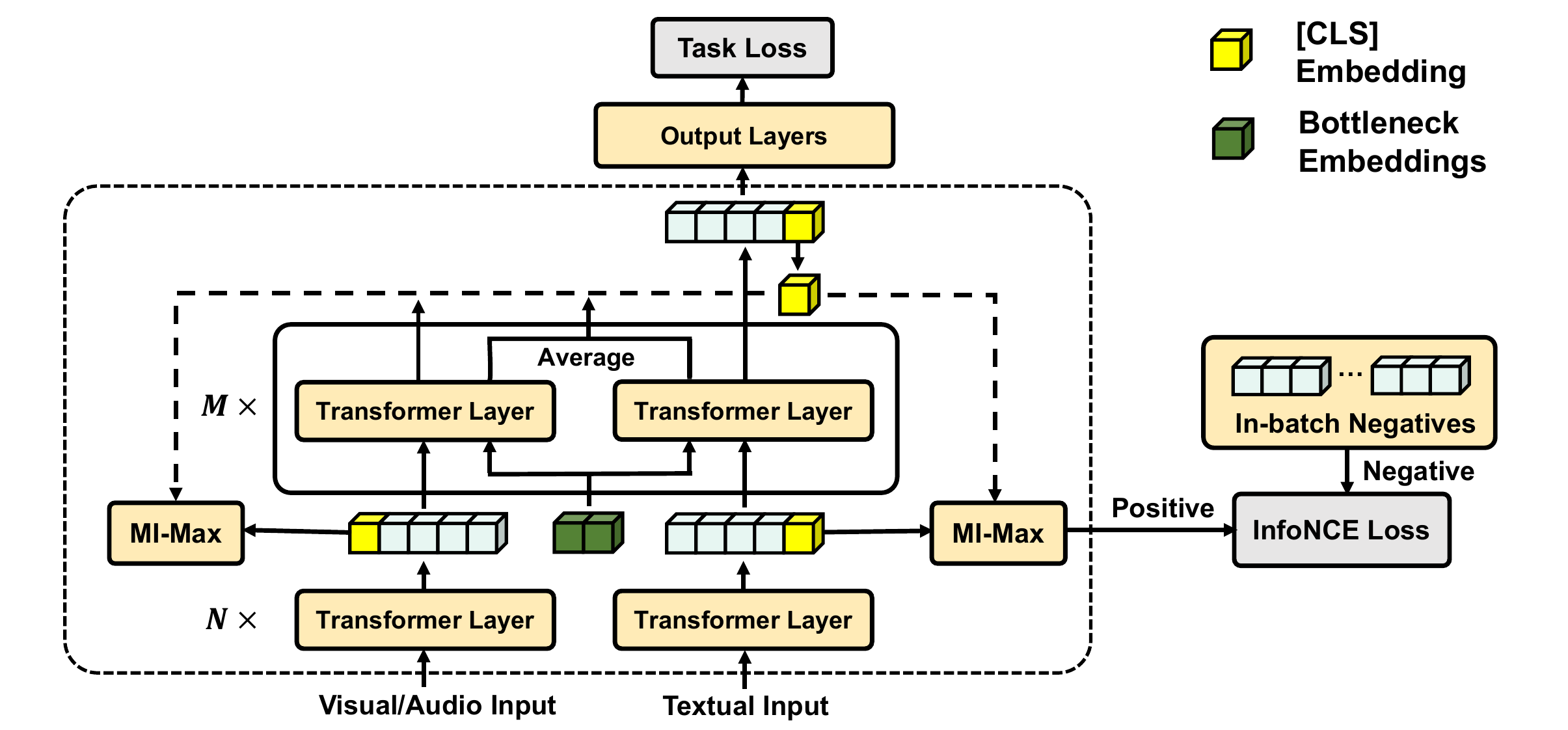}
    \caption{Overview of our denoising fusion bottleneck (DBF) model. It consists of $N$ Transformer layers to encode videos and texts, and $M$ Transformer layers with fusion bottlenecks for multimodal fusion. We incorporate a mutual information maximization (MI-Max) InfoNCE loss to regulate the bottleneck module, aiming to preserve key information in both modalities from being filtered.
    }
    \label{fig:overall structure}	
\end{figure*}

\section{Methodology}
Our denoising fusion bottleneck (DBF) model aims to fuse multimodal inputs from videos to make a comprehensive prediction. 
The overall architecture of DBF is shown in Figure \ref{fig:overall structure}.
It first employs a fusion bottleneck module with a restrained receptive field to filter out noise and redundancy when fusing different modalities in videos. 
Then, DBF maximizes mutual information between fusion results and unimodal inputs to supervise the learning of the fusion bottleneck, aiming to preserve more representative information in fusion results.

\subsection{Problem Definition}
In video multimodal fusion tasks, for each video, the input comprises three sequences of encoded features from textual ($t$), visual ($v$), and acoustic ($a$) modalities.
These input features are represented as $X_m \in \mathbb{R}^ {l_m \times d_m}$,  where $m \in \{t,v,a\}$, and $l_m$ and $d_m$ denote the sequence length and feature dimension for modality $m$, respectively.
The goal of DBF is to extract and integrate task-related information from these input representations to form a unified fusion result $Z \in \mathbb{R}^ {l \times d}$. 
In this paper, we evaluate the quality of the fusion result $Z$ on two tasks: video multimodal sentiment analysis and video multimodal summarization. 

For sentiment analysis, we utilize $Z$ to predict the emotional orientation of a video as a discrete category $\hat{y}$ from a predefined set of candidates $\mathcal{C}$
\begin{equation}
    \hat{y} = \operatorname{argmax}_{y_j \in \mathcal{C}} \operatorname{P}_{\Theta}(y_j \mid Z), 
\end{equation} 
or as a continuous intensity score $\hat{y} \in \mathbb{R}$
\begin{equation}
    \hat{y} = \operatorname{P}_{\Theta}(Z), 
\end{equation}
where $\Theta$ denotes the model parameters. 

For summarization, we generate a summary sequence $\hat{S} = (s_1, ..., s_l)$ based on $Z$: 
\begin{equation}
\label{eq1}
\hat{S} = \text{argmax}_{S} \operatorname{P}_{\Theta}(S \mid Z).
\end{equation}

\subsection{Fusion Bottleneck}

As shown in Figure \ref{fig:overall structure}, we first employ a fusion bottleneck with a restrained receptive field to perform multimodal fusion and filter out noise and redundancy in videos.
Specifically, fusion bottleneck forces cross-modal information flow passes via randomly initialized bottleneck embeddings $B \in \mathbb{R}^{{l_b \times d_m}}$ with a small sequence length, where $d_m$ denotes dimension of features and $l_b \ll l$. 
The restrained receptive field of $B$ forces model to collate and condense unimodal information before sharing it with the other modalities.

With a small length $l_b$, embedding $B$ acts like a bottleneck in cross-modal interaction. 
In the fusion bottleneck module, unimodal features cannot directly attend to each other and they can only attend to the bottleneck embeddings $B$ to exchange information in it. 
Meanwhile, the bottleneck can attend to all of the modalities, which makes information flow across modalities must pass through the bottleneck with a restrained receptive field. 
The fusion bottleneck module forces the model to condense and collate information and filter out noise and redundancy.

Specifically, in the fusion bottleneck module, with bottleneck embeddings $B$ and unimodal features $X_m$, the fusion result is calculated as follows:
\begin{equation}
\label{eq:trans}
[X_m^{l+1}||B_m^{l+1}] = \text{Transformer}([X_m^{l}||B^{l}]), 
\end{equation}
\begin{equation}
\label{eq:avg}
B^{l+1} = \text{Mean}(B_m^{l+1}), 
\end{equation} 
where $l$ denotes the layer number and $||$ denotes the concatenation operation.
As shown in Equation \ref{eq:trans} and \ref{eq:avg}, each time a Transformer layer is passed, bottleneck embedding $B$ is updated by unimodal features.
In turn, unimodal features integrate condensed information from other modalities through bottleneck embeddings $B$.
Finally, we output the text features $X_t^{L}$ of the last layer $L$, which are injected with condensed visual and audio information, as the fusion result.


\subsection{Fusion Mutual Information Maximization}

The fusion bottleneck module constrains information flow across modalities in order to filter out noise and redundancy. 
However, it may result in loss of critical information as well when fusion bottleneck selects what information to be shared.
To alleviate this issue, we employ a mutual information maximization (MI-Max) module to preserve representative and salient information from redundant modalities in fusion results. 

Mutual information is a concept from information theory that estimates the relationship between pairs of variables.
Through prompting the mutual information between fusion results $Z$ and multimodal inputs $X_m$, we can capture modality-invariant cues among modalities \citep{han2021improving} and keep key information preserved by regulating the fusion bottleneck module. 


Since direct maximization of mutual information for continuous and high-dimensional variables is intractable \citep{belghazi2018mine}, we instead minimize the lower bound of mutual information as \citet{han2021improving} and \citet{oord2018representation}.
To be specific, we first construct an opposite path from $Z$ to predict $X_m$ by an MLP $\mathcal{F}$. 
Then, to gauge correlation between the prediction and $X_m$, we use a normalized similarity function as follows:
\begin{equation}
\text{sim}(X_m, Z) = 
\text{exp}
\left(
\frac{X_m}{ \left\| X_m \right\| ^{2} } 
\odot
\frac{\mathcal{F}(Z)}{\left\| \mathcal{F}(Z) \right\| ^{2}} 
\right),
\end{equation} 
where $\mathcal{F}$ generates a prediction of $X_m$ from $Z$, $\|\cdot\|^{2}$ is the Euclidean norm, and $\odot$ denotes element-wise product. 
Then, we incorporate this similarity function into the noise-contrastive estimation framework \citep{gutmann2010noise} and produce an InfoNCE loss \citep{oord2018representation} which reflects the lower bound of the mutual information:
\begin{equation}
    \mathcal{L}_{\text{NCE}}^{z, m} =
    -\mathbb{E}_{X_{m},Z}\left
    [
    \log \frac
    {e^{\operatorname{sim}\left(x_{m}^{\small{+}},\mathcal{F}(Z) \right) }}
    {\sum_{k=1}^{K} 
    e^{\operatorname{sim}\left(\tilde{x}_{m}^{k}, \mathcal{F}(Z) \right)}}\right
    ]
\end{equation}
where $\tilde{x}_m=\left\{\tilde{x}^{1}, \ldots, \tilde{x}^{K}\right\}$ is the negative unimodal inputs that are not matched to the fusion result $Z$ in same batch. 
Finally, we compute loss for all modalities as follows:
\begin{equation}
\mathcal{L}_{\text{NCE}} = \alpha (\mathcal{L}_{\text{NCE}}^{z,v} + \mathcal{L}_{\text{NCE}}^{z,a} + \mathcal{L}_{\text{NCE}}^{z,t})   
\end{equation}
where $\alpha$ is a hyper-parameter that controls the impact of MI-Max.

By minimizing $\mathcal{L}_{\text{NCE}}$, on the one hand, we maximize the lower bound of the mutual information between fusion results and unimodal inputs; on the other hand, we encourage fusion results to reversely predict unimodal inputs as well as possible, which prompts retaining of representative and key information from different modalities in fusion results.

\begin{table*}[t]
\centering
\setlength{\tabcolsep}{7pt}
\begin{tabular}{lcccccc}
\toprule
\multicolumn{1}{l|}{\multirow{2}{*}{\textbf{Method}}} & \multicolumn{5}{c}{\textbf{MOSI}}  \\
\multicolumn{1}{c|}{} & \textbf{MAE($\downarrow$)}   & \textbf{Corr($\uparrow$)}  & \textbf{Acc-7($\uparrow$)}  & \textbf{Acc-2($\uparrow$)}   & \textbf{F1($\uparrow$)}     \\
\midrule
\multicolumn{1}{l|}{MulT \citep{mult} } 			&0.871 &0.698 	&40.0	&- / 83.0 		&- / 82.8 		 \\
\multicolumn{1}{l|}{TFN \citep{zadeh2017tensor}}    			&0.901 &0.698 	&34.9	&- / 80.8 		&- / 80.7 		 \\
\multicolumn{1}{l|}{LMF \citep{lmf}}  			&0.917 &0.695 	&33.2	&- / 82.5 		&- / 82.4 		 \\
\multicolumn{1}{l|}{MFM \citep{mfm}}  			&0.877 &0.706 	&35.4	&- / 81.7 		&- / 81.6 		 \\
\multicolumn{1}{l|}{ICCN \citep{iccn}}  			&0.860 &0.710 	&39.0	&- / 83.0 		&- / 83.0 		 \\
\multicolumn{1}{l|}{MISA \citep{hazarika2020misa}}  			&0.783 &0.761 	&42.3	&81.8 / 83.4   	&81.7 / 83.6   	\\
\multicolumn{1}{l|}{Self-MM \citep{yu2021learning}} 	  	 	& 0.712 & 0.795 & 45.8 & 82.5 / 84.8	& 82.7 / 84.9		\\
\multicolumn{1}{l|}{MMIM$^{\dag}$ \citep{han2021improving}} 	  	 	& 0.700 & \textbf{0.800} & \textbf{46.7} & 84.2 / 86.1 	& 84.0 / 86.0	  	\\
\midrule
\multicolumn{1}{l|}{DBF} 	 		& \textbf{0.693} & \textbf{0.801} & 44.8 & \textbf{85.1 / 86.9} 	& \textbf{85.1 / 86.9}		\\
\bottomrule
\end{tabular}
\caption{Results of multimodal sentiment analysis on MOSI. $\dag$ indicates the previous state-of-the-art model.}
\label{tab:mosi}
\end{table*}

\begin{table*}[t]
\centering
\setlength{\tabcolsep}{7pt}
\begin{tabular}{l | cccccccccc}
\toprule
\multicolumn{1}{l|}{\multirow{2}{*}{\textbf{Method}}} & \multicolumn{5}{c}{\textbf{MOSEI}} \\
\multicolumn{1}{c|}{} & \textbf{MAE($\downarrow$)}   & \textbf{Corr($\uparrow$)}  & \textbf{Acc-7($\uparrow$)}  & \textbf{Acc-2($\uparrow$)}     & \textbf{F1($\uparrow$)}     \\
\midrule
MulT \citep{mult}    	&0.580 &0.703  &51.8   &- / 82.3   &- / 82.5 		 		 \\
TFN \citep{zadeh2017tensor} 		&0.593 &0.700     &50.2   &- / 82.1   &- / 82.5 		 		 \\
LMF \citep{lmf}  	&0.677 &0.695     &48.0   &- / 82.1   &- / 82.0 		 		 \\
MFM \citep{mfm}  	&0.717 &0.706     &51.3   &- / 84.3   &- / 84.4 		 		 \\
ICCN \citep{iccn}  	&0.565 &0.713    &51.6   &- / 84.2   &- / 84.2 		 		 \\
MISA \citep{hazarika2020misa}  	&0.555 &0.756    &52.2  &83.8 / 85.3 &83.6 / 85.5   	   	 \\
Self-MM \citep{yu2021learning}  & 0.529 & 0.767 & 53.5 & 82.7 / 85.0 & 83.0 / 84.9   \\
MMIM$^{\dag}$ \citep{han2021improving}  & 0.526 & \textbf{0.772} & \textbf{54.2} & 82.2 / 86.0 & 82.7 / 85.9   \\
\midrule
DBF  & \textbf{0.523} & \textbf{0.772}      & \textbf{54.2}      &  \textbf{84.3 / 86.4}           & \textbf{84.8 / 86.2}  \\
\bottomrule
\end{tabular}
\caption{Results of multimodal sentiment analysis on MOSEI. $\dag$ indicates the previous state-of-the-art model. }
\label{tab:mosei}
\end{table*}

\section{Experiments}

\subsection{Tasks, Datasets, and Metrics}

We evaluate fusion results of DBF on two video multimodal tasks: video multimodal sentiment analysis and video multimodal summarization.

\paragraph{Video Multimodal Sentiment Analysis} 
Video multimodal sentiment analysis is a regression task that aims to collect and tackle data from multiple resources (text, vision and acoustic) to comprehend varied human emotions. 
We do this task on MOSI \citep{zadeh2016mosi} and MOSEI \citep{mosei} datasets.
The MOSI dataset contains 2198 subjective utterance-video segments, which are manually annotated with a continuous opinion score between [-3, 3], where -3/+3 represents strongly negative/positive sentiments. 
The MOSEI dataset is an improvement over MOSI, which contains 23453 annotated video segments (utterances), from 5000 videos, 1000 distinct speakers and 250 different topics. 

Following \citep{hazarika2020misa}, we use the same metric set to evaluate sentiment intensity predictions: 
MAE (mean absolute error), which is the average of absolute difference value between predictions and labels;
Corr (Pearson correlation) that measures the degree of prediction skew; 
Acc-7 (seven-class classification accuracy) ranging from -3 to 3; 
Acc-2 (binary classification accuracy) and F1 score computed for positive/negative and non-negative/negative classification results.

\paragraph{Video Multimodal Summarization} 
The summary task aims to generate abstractive summarization with videos and their corresponding transcripts.
We set How2 dataset \citep{sanabria2018how2} as benchmark for this task, which is a large-scale dataset consists of 79,114 short instructional videos, and each video is accompanied by a human-generated transcript and a short text summary.

Following \citep{yu2021vision}, to evaluate summarization, we use metrics as follows:
ROUGE \citep{rouge} (ROUGE-{1, 2, L}) and BLEU \citep{bleu} (BLEU-{1, 2, 3, 4}), which calculate the recall and precision of n-gram overlaps, respectively; 
METEOR \citep{meteor}, which evaluates matching degree of word stems, synonyms and paraphrases;
CIDEr \citep{cider} is an image captioning metric to compute the cosine similarity between TF-IDF weighted n-grams.
 
\subsection{Experimental Settings}


For sentiment analysis task, we use BERT-base~\citep{devlin2018bert} to encode text input and extract the [CLS] embedding from the last layer.
For acoustic and vision, we use COVAREP \citep{degottex2014covarep} and Facet \footnote{https://imotions.com/platform/} to extract audio and facial expression features. 
The visual feature dimensions are 47 for MOSI, 35 for MOSEI, and the audio feature dimensions are 74 for both MOSI and MOSEI.

For summarization, we use BART \citep{lewis2019bart} as the feature extractor and inject visual information in the last layer of the BART encoder.
For vision, a 2048-dimensional feature representation is extracted for every 16 non-overlapping frames using a 3D ResNeXt-101 model \citep{hara2018can}, which is pre-trained on the Kinetics dataset \citep{kay2017kinetics}.
Details of the hyper-parameters are given in Appendix~\ref{sec:app1}.
For frameworks and hardware, we use the deep learning framework PyTorch \citep{2017Automatic} and Huggingface \footnote{https://huggingface.co/} to implement our code. We use a single Nvidia GeForce A40 GPU for sentiment analysis experiments and two for summarization.

\subsection{Overall Results}
\label{sec:results}

\begin{table*}[t]
\setlength{\tabcolsep}{4pt}
\centering
\begin{tabular}{l|ccccccccc}
\toprule
 \multirow{2}{*}{\textbf{Method}} & \multicolumn{9}{c}{\textbf{How2}}  \\
   & \textbf{R-1}  & \textbf{R-2}  & \textbf{R-L}  & \textbf{B-1}  & \textbf{B-2}  & \textbf{B-3}  & \textbf{B-4}  & \textbf{METEOR} & \textbf{CIDEr} \\
\midrule
 HA (RNN) \citep{han}	& 60.3	& 42.5	& 55.7	& 57.2	& 47.7	& 41.8	& 37.5	& 28.8	& 2.48	\\
 HA (TF) \citep{han}	& 60.2	& 43.1	& 55.9	& 58.6	& 48.3	& 43.3	& 38.1	& 28.9	& 2.51	\\
 MFFG (RNN) \citep{liu2020multistage} 	& 62.3	& 46.1	& 58.2	& 59.1	& 50.4	& 45.1	& 41.1	& 30.1	& 2.69	\\
 MFFG (TF) \citep{liu2020multistage} 	& 61.6	& 45.1          & 57.4          & 60.0          & 50.9          & 45.3          & 41.3          & 29.9	& 2.67	\\
 VG-GPLMs$^{\dag}$ \citep{yu2021vision}	& 68.0	& 51.4	& 63.3	& 65.2          & 56.3          & 50.4          & 46.0          & 34.0            & 3.28	\\
 \midrule
 DBF	& \textbf{70.1} & \textbf{54.7} & \textbf{66.0} & \textbf{67.2} & \textbf{58.9} & \textbf{53.3} & \textbf{49.0} & \textbf{35.5}   & \textbf{3.56} \\
\bottomrule
\end{tabular}
\caption{Results of multimodal summarization task on How2. The $\dag$ indicates the previous state-of-the-art model. We denote ROUGE and BLEU by R and B respectively.}
\label{tab:how2}
\end{table*}

\begin{table*}[t]
\centering
\setlength{\tabcolsep}{15pt}
\begin{tabular}{l|cc|cc}
\toprule
\multirow{2}{*}{\textbf{Model}} & \multicolumn{2}{c|}{\textbf{MOSI}} & \multicolumn{2}{c}{\textbf{MOSEI}} \\
   & \textbf{MAE ($\downarrow$)}  & \textbf{F1 ($\uparrow$)}  & \textbf{MAE ($\downarrow$)}  & \textbf{F1 ($\uparrow$)}     \\
\midrule
1) Ours                                   & \textbf{0.693}      &\textbf{85.07 / 86.88}       & \textbf{0.523}       & \textbf{84.78 / 86.19}       \\
2) (-) MI-Max   &0.697            &83.08 / 85.28             & 0.536        & 80.94 / 85.58       \\
3) (-) bottleneck                         & 0.750      & 82.84 / 83.63       & 0.537       & 77.52 / 83.81       \\
\midrule
4) (-) Language $l$                       & 1.391      & 55.54 / 54.95       & 0.817             & 67.63 / 64.01            \\
5) (-) Visual $v$                         & 0.700      & 82.78 / 84.33       & 0.541       & 78.42 / 84.05      \\
6) (-) Audio $a$                          & 0.720      & 83.02 / 85.86       & 0.536       & 80.22 / 85.02      \\
\midrule
7) Visual-based                           & 1.372      & 57.06 / 57.83       &  0.536           &  83.41 / 85.47           \\
8) Audio-based                            & 1.194      & 67.95 / 70.49       &0.537             &  83.80 / 85.76           \\
\bottomrule
\end{tabular}
\caption{Results of ablation study. (-) represents removal for the mentioned factors. 
Model 1 represents the best performing model in each dataset; Model 2,3 presents the effect of MI module and bottleneck module; Model 4,5,6 depicts the effect of individual modalities; Model 7,8 presents the variants of our model as defined in Section \ref{sec:ablation}.}
\label{tab:ablation}
\end{table*}

\begin{figure*}[t]
    \centering
    \includegraphics[width=0.8\linewidth]{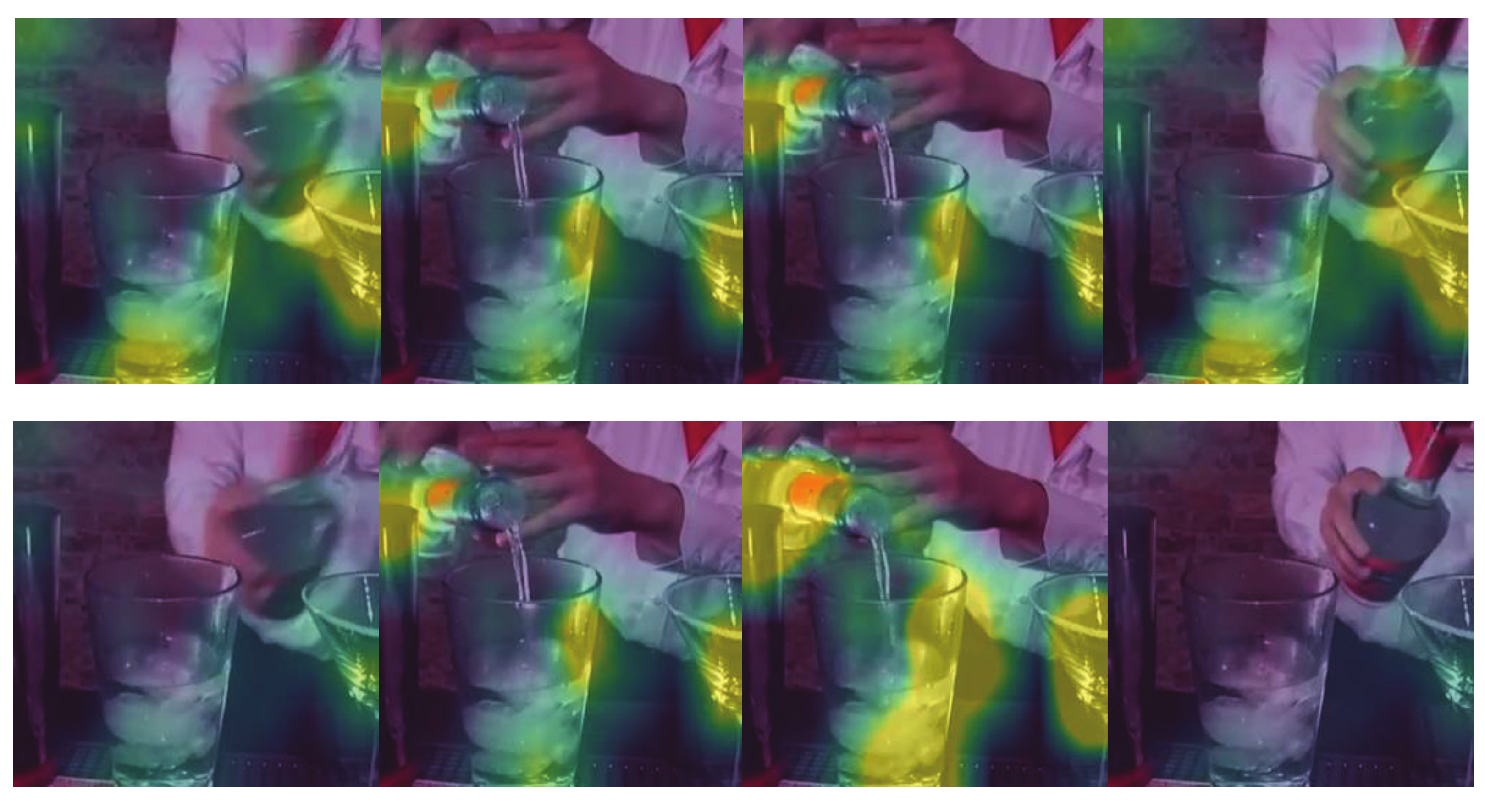}
    \caption{Comparison of Grad-CAM visualizations of baseline method VG-GPLMs \citep{yu2021vision} (top) and DBF (bottom).
    In contrast to even attention to different frames of the baseline method, DBF ignores redundancy and noise in consecutive frames and highly focuses on the key information (\emph{pouring wine} in this example) in a particular frame. }
    \label{fig:gradcam}	
\end{figure*}

We compare performance against DBF by considering various baselines as below: 
For multimodal sentiment analysis, we compare with 
MulT	\citep{mult}, 
TFN	\citep{zadeh2017tensor},
LMF	\citep{lmf},
MFM	\citep{mfm},
ICCN	\citep{iccn},
MISA	\citep{hazarika2020misa},
Self-MM	\citep{yu2021learning} and
MMIM	\citep{han2021improving}.
For multimodal summarization, we compare with 
HA \citep{han}
MFFG \citep{liu2020multistage} 
VG-GPLMs \citep{yu2021vision}.
Details of baselines are in Appendix~\ref{sec:app2}.
The comparative results for sentiment analysis are presented in Table \ref{tab:mosi} (MOSI) and Table \ref{tab:mosei} (MOSEI). 
Results for summarization are presented in Table \ref{tab:how2} (How2).

We find that DBF yields better or comparable results to state-of-the-art methods. 
To elaborate, DBF significantly outperforms state-of-the-art in all metrics on How2 and in most of metrics on MOSI and MOSEI. 
For other metrics, DBF achieves very closed performance to state-of-the-art. 
These outcomes preliminarily demonstrate the efficacy of our method in video multimodal fusion.

From the results, we can observe that our model achieves more significant performance improvement on summary task than sentiment analysis.
There could be two reasons for this: 1) the size of two datasets is small, yet DBF requires a sufficient amount of data to learn noise and redundancy patterns for this type of video.
2) Visual features are extracted by Facet on sentiment analysis task and more 3D ResNeXt-101 on summary task respectively.
Compared to sentiment analysis task, summary task employ a more advanced visual extractor and DBF is heavily influenced by the quality of visual features.

\subsection{Ablation Study}
\label{sec:ablation}

\paragraph{Effect of Fusion Bottleneck and MI-Max}
As shown in Table \ref{tab:ablation}, we first remove respectively MI-Max module and exchange fusion bottleneck module with vanilla fusion methods to observe the effects on performance.
We observe that fusion bottleneck and MI-Max both help better fusion results, and the combination of them further improves performance, which reflects the necessity of removing noise while maintaining representative information.

\paragraph{Effect of Modalities}
Then we remove one modality at a time to observe the effect on performance. 
Firstly, we observe that the multimodal combination provides the best performance, indicating that our model can learn complementary information from different modalities.
Next, we observe that the performance drops sharply when the language modality is removed.
This may be due to the fact that text has higher information density compared to redundant audio and visual modalities.
It verifies two things: 1) It is critical to remove noise and redundancy to increase information density of visual and audio modalities when doing fusion. 2) Text-centric fusion results may help improve performance on multimodal summary and sentiment analysis tasks.

\paragraph{Effect of Center Modality}
As mentioned above, text-centric fusion results tend to perform better as low information intensity and high redundancy in other modalities. Thus, we evaluate fusion results based on acoustic and vision modality respectively on downstream tasks.
We observe an obvious decline in performance when audio or visual modality is used as the central modality.

\subsection{Case Study}

In this section, 
we first calculate standard deviation and normalized entropy over visual attention scores in the Grad-CAM heatmaps \citep{selvaraju2017grad} for DBF and baseline method VG-GPLMs \citep{yu2021vision} respectively.
These two metrics show the sharpness of visual attention scores, indicating whether the model focuses more on key frames and ignores redundant content.
Then, we compute visualizations on Grad-CAM heatmaps acquired before to show the ability of DBF to filter out redundancy and preserve key information.


\paragraph{Statistics of Visualization Results}
Grad-CAM is a visualization method of images, it obtains visualization heatmaps by calculating weights and gradients during backpropagation, and in this paper we extend Grad-CAM to videos.
Further, to quantify this sharpness of visual attention, we calculate standard deviation and normalized entropy on Grad-CAM heatmaps over the test split on How2 dataset. 
For results, DBF gets 0.830, 0.008, baseline gets 0.404, 0.062 in deviation and normalized entropy respectively.
DBF holds a higher deviation and lower entropy, which indicates sharper visual attention maps to discriminate redundancy and key frames.

\paragraph{Visualization Example}

Figure \ref{fig:gradcam} provides Grad-CAM visualizations of DBF and baseline method. 
As we can see, DBF has more sharp attention over continuous frames and ignores redundancy while preserving critical information in visual inputs. 




\section{Conclusion}
\label{sec:conclution}
In this paper, we propose a denoising video multimodal fusion system DBF which contains a fusion bottleneck to filter out redundancy with noise,
a mutual information module to preserve key information in fusion results.
Our model alleviates redundancy and nosie problem in video multimodal fusion and makes full use of all representative information in redundant modalities (vision and acoustic).
In the experiments, we show that our model significantly and consistently outperforms state-of-the-art video multimodal models.
In addition, we demonstrate that DBF can appropriately select necessary contents and neglect redundancy in video by comprehensive ablation and visualization studies.

In the future, we will explore the following directions:
(1) We will try to extend the proposed DBF model to more multimodal fusion tasks such as humor detection.
(2) We will incorporate vision-text pretraining backbones into our DBF model to further improve its performance.

\section*{Limitations}
First, limited by the category of video multimodal fusion tasks, we do not perform experiments on more tasks to better validate the effectiveness of our method, and we hope to extend our model to more various and complete benchmarks in future work.
Secondly, as shown in Section~\ref{sec:results}, our model achieves relatively slight performance improvement on sentiment analysis task.
For reasons, our model may be dependent on the scale of datasets to learn noise and redundancy patterns in video, which needs to be further improved and studied.

\section*{Acknowledgement}

This paper is supported by the National Key Research and Development Program of China 2020AAA0106700 and NSFC project U19A2065.

\bibliography{anthology,custom}
\bibliographystyle{acl_natbib}

\clearpage
\appendix

\section*{Appendix}

\section{Hyper-parameters}
\label{sec:app1}
We set hyper-parameters as shown in Table \ref{tab:hyper} for best performance.
For optimization, we utilize the Adam optimizer with warmup. The training duration of each model is governed by early-stopping strategy with a patience of 10 epochs. 
\begin{table}[h]
\centering
\setlength{\tabcolsep}{1.5pt}
\begin{tabular}{@{}lccc@{}}
\toprule
\textbf{Hyper-Parameter}                      & \textbf{MOSI} & \textbf{MOSEI} & \textbf{How2}     \\
\midrule
Batch size                & 32       & 96        & 80       \\
Bottleneck length         & 2        & 4         & 8        \\
Num of bottleneck layers          & 4        & 4         & 4        \\
$\alpha$                  & 0.05     & 0.1       & 0.1      \\
Learning rate $\eta_{\text{DBF}}$       & 2e-05 & 2e-03  & 3e-04 \\
Learning rate $\eta_{\text{Backbone}}$ & 1e-04 & 5e-05  & 6e-05 \\
Fusion size        & 128      & 128       & 768     \\
\bottomrule
\end{tabular}
\caption
{Hyper-parameters for the best performance. 
$\eta_{\text{Backbone}}$ denotes the learning rate of parameters of the backbone pretrained model. 
$\eta_{\text{DBF}}$ denotes the learning rate of new parameters introduced by our DBF model.
}
\label{tab:hyper}
\end{table}

\section{Baselines}
\label{sec:app2}

For multimodal sentiment analysis:
\paragraph{MulT \citep{mult} :}
a multimodal transformer architecture model with directional pairwise cross-attention, which translates one modality to another.
\paragraph{TFN \citep{zadeh2017tensor}}
based on tensor outer product to capture multiple-modal interactions.
\paragraph{LMF \citep{lmf} :}
an advanced version of TFN model.
\paragraph{MFM \citep{mfm} :}
a model that factorizes representations into two sets of independent factors: multimodal discriminative and modality-specific generative factors. 
\paragraph{ICCN \citep{iccn} :}
an adversarial encoder-decoder classifier framework-based model to learn a modality-invariant embedding space.
\paragraph{MISA \citep{hazarika2020misa} }
 projects each modality to two distinct subspaces.
\paragraph{Self-MM \citep{yu2021learning} }
propose a label generation module based on the self-supervised learning strategy to acquire independent unimodal supervision.
\paragraph{MMIM \citep{han2021improving} }
hierarchically maximizes the mutual information in unimodal input pairs and between multimodal fusion result and unimodal input.

For multimodal summarization, We compare DBF with the following baselines:
\paragraph{HA \citep{han} :}
a sequence-to-sequence multimodal fusion model with hierarchical attention.

\paragraph{MFFG \citep{liu2020multistage} :}
a multistage fusion network with the fusion forget gate module, which controls the flow of redundant information between multimodal long sequences via a forgetting module.

\paragraph{VG-GPLMs \citep{yu2021vision} :}
a BART-based and vision guided model for multimodal summarization task, which use attention-based add-on layers to incorporate visual information.

\end{document}